\documentclass{article}

\usepackage{microtype}
\usepackage{graphicx}
\usepackage{subfigure}
\usepackage{booktabs} 

\usepackage{hyperref}

\usepackage[accepted]{icml2024}

\usepackage{amsmath}
\usepackage{amssymb}
\usepackage{mathtools}
\usepackage{amsthm}

\usepackage[capitalize,noabbrev]{cleveref}

\usepackage{caption}
\usepackage{listings}
\usepackage{color}

\definecolor{codegreen}{rgb}{0,0.6,0}
\definecolor{codegray}{rgb}{0.5,0.5,0.5}
\definecolor{codepurple}{rgb}{0.58,0,0.82}
\definecolor{backcolour}{rgb}{0.95,0.95,0.92}

\lstdefinestyle{mystyle}{
    backgroundcolor=\color{backcolour},   
    commentstyle=\color{codegreen},
    keywordstyle=\color{magenta},
    numberstyle=\tiny\color{codegray},
    stringstyle=\color{codepurple},
    basicstyle=\footnotesize,
    breakatwhitespace=false,         
    breaklines=true,                 
    captionpos=b,                    
    keepspaces=true,                 
    numbers=left,                    
    numbersep=5pt,                  
    showspaces=false,                
    showstringspaces=false,
    showtabs=false,                  
    tabsize=2
}

\usepackage{dsfont}
\usepackage{soul}
\usepackage{enumitem}

\theoremstyle{plain}

\theoremstyle{definition}

\theoremstyle{remark}

\usepackage[textsize=tiny]{todonotes}

\icmltitlerunning{Learning Memory Mechanisms for Decision Making through Demonstration}

\begin{document}

\twocolumn[
\icmltitle{Learning Memory Mechanisms for\\
Decision Making through Demonstrations}



\icmlsetsymbol{equal}{*}

\begin{icmlauthorlist}
\icmlauthor{William Yue}{ut}
\icmlauthor{Bo Liu}{ut}
\icmlauthor{Peter Stone}{ut,sony}
\end{icmlauthorlist}

\icmlaffiliation{ut}{The University of Texas at Austin}
\icmlaffiliation{sony}{Sony AI}

\icmlcorrespondingauthor{William Yue}{william.yue@utexas.edu}

\icmlkeywords{Machine Learning, ICML, Decision-Making, Memory, Learning from Demonstration, Imitation Learning, Reinforcement Learning, Transformers, Behavioral Cloning, Attention}

\vskip 0.3in
]



\printAffiliationsAndNotice{}  

\begin{abstract}
In Partially Observable Markov Decision Processes, integrating an agent's history into memory poses a significant challenge for decision-making. Traditional imitation learning, relying on observation-action pairs for expert demonstrations, fails to capture the expert's memory mechanisms used in decision-making. To capture memory processes as demonstrations, we introduce the concept of \textbf{memory dependency pairs} $(p, q)$ indicating that events at time $p$ are recalled for decision-making at time $q$. We introduce \textbf{AttentionTuner} to leverage memory dependency pairs in Transformers and find significant improvements across several tasks compared to standard Transformers when evaluated on Memory Gym and the Long-term Memory Benchmark. Code is available at \href{https://github.com/WilliamYue37/AttentionTuner}{https://github.com/WilliamYue37/AttentionTuner}.
\end{abstract}

\begin{figure*}[ht]

\begin{center}
\centerline{\includegraphics[width=\textwidth]{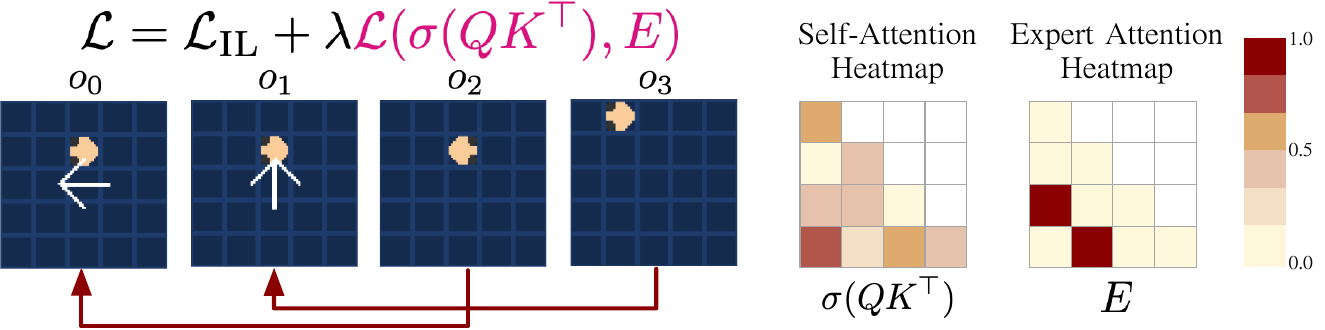}}
\caption{The red arrows indicate episodic memory dependencies labeled by an expert. The correct action to take in $o_2$ depends on $o_0$ and the correct action to take in $o_3$ depends on $o_1$. These memory dependency pairs are used to create the expert self-attention matrix $E \in \{0,1\}^{n \times n}$ where $n$ is the length of the sequence and $E_{ij} = 1$ only if the expert has indicated that observation $o_j$ should be recalled from memory at observation $o_i$, and $E_{ij} = 0$ otherwise. A binary cross entropy loss is taken between $E$ and the learner's self-attention matrix $\sigma(QK^\top)$ to form the memory loss $\mathcal{L}(\sigma(QK^\top), E)$ that encourages the learner to learn the expert's memory mechanism. The memory loss is scaled by $\lambda$ to match the magnitude of $\mathcal{L}_{\mathrm{IL}}$ and then added to form the final loss used during training.}
\label{fig1}
\end{center}

\end{figure*}

\section{Introduction}
\label{intro}

Partially Observable Markov Decision Processes (POMDPs) offer a framework for modeling decision-making in environments where the agent's information is incomplete, a common situation in real-world scenarios such as a robot operating based on limited camera observations. Making effective decisions under such conditions necessitates incorporating the agent's history, which can be encoded through memory mechanisms like Recurrent Neural Networks (RNNs) \cite{hausknecht2017deep, karkus2017qmdpnet, zhu2018improving, igl2018deep, hafner2019learning} or self-attention architectures such as Transformers \cite{esslinger2022deep}. 

However, it’s not always straightforward for a fully automated system to identify which points in history are crucial to remember for a particular decision. On the other hand, when humans learn, we are often taught not just the actions we need to take at the moment, but also which past events and memories should be recalled. For instance, a survival instructor might instruct students to recollect previously observed landmarks for navigating back to base camp or a coach could ask an athlete to recall a past encounter with an opponent when making their next move.

With this motivation in mind, this study concentrates on learning memory mechanisms essential for decision-making in POMDP tasks via expert demonstrations, also known as imitation learning. Standard imitation learning methods, which involve experts providing observation-action pairs, are insufficient in POMDP tasks as they do not capture the memory processes experts employ during decision-making.

To capture memory mechanisms through demonstration, we introduce the use of \textbf{memory dependency pairs} $(p, q)$, where $p < q$, indicating that the observation at time $p$ ought to be recalled for decision-making at time $q$. These memory dependency pairs can be integrated into the widely-used Transformers \cite{vaswani2023attention} by applying a loss to the self-attention matrix to reinforce attention between tokens representing times $p$ and $q$ (Figure~\ref{fig1}). 

The main contributions of this paper are as follows:
\begin{itemize}
    \item We introduce \textbf{memory dependency pairs} to incorporate memory mechanisms into demonstrations for imitation learning in POMDPs and to improve long-term credit assignment.
    \item We develop the \textbf{Long-term Memory Benchmark} (LTMB) for evaluating long-term memory capabilities in decision-making tasks.
    \item We introduce \textbf{AttentionTuner}, a novel method for leveraging memory dependency pairs in self-attention architectures, and benchmark it against vanilla Transformers on Memory Gym \cite{pleines2024memory} and LTMB. Empirical analyses show that AttentionTuner significantly improves success rates on four tasks and aids the optimizer in consistently navigating the loss landscape towards solutions with better generalizability compared to those found by optimizing the vanilla Transformer. Ablations reveal that these improvements in learning can be attained with as few as 0.1\% of demonstrations annotated.
\end{itemize}

\section{Background}
\label{background}

This section defines the notation and framework for imitation learning in partially observable environments and provides a concise overview of Transformer architectures. This notation will be used to define AttentionTuner in Section~\ref{method}.

\subsection{Imitation Learning in Partially Observable Environments}
\label{IL_intro}

Imitation learning algorithms aim to learn a policy $\pi_\theta$ parameterized by $\theta$ by imitating a set of expert demonstrations $D = \{\tau_i\}_{i = 1 \ldots M}$. Each demonstration trajectory $\tau_i$ is a sequence of observation-action pairs ${(o_j, a_j)},~~j = 1 \ldots |\tau_i|$, where $|\tau_i|$ denotes the trajectory length. These trajectories are generated from a Partially Observable Markov Decision Process (POMDP), which is characterized by the tuple $\langle \mathcal{S}, \mathcal{O}, \mathcal{A}, T, O, \rho_0 \rangle$. Here, $\mathcal{S}$ represents the state space, $\mathcal{O}$ the observation space, $\mathcal{A}$ the action space, $T: \mathcal{S} \times \mathcal{A} \times \mathcal{S} \to [0, 1]$ the transition dynamics, $O: \mathcal{S} \times \mathcal{O} \to [0, 1]$ the observation function, and $\rho_0$ the initial state distribution. There are several approaches to imitation learning, including offline methods that do not require environmental interactions like behavioral cloning~\citep{schaal1999imitation}, online methods like Generative Adversarial Imitation Learning (GAIL) \cite{ho2016generative} and inverse reinforcement learning methods \cite{ng2000inverse}. In this study, we focus on behavioral cloning, where the objective is to minimize the negative log-likelihood loss function for a discrete action space:
\begin{equation}
\label{ILloss}
\mathcal{L}_{\text{IL}}(\theta) = -\mathbb{E}_{(o,y) \sim D}\bigg[\sum_a^A \mathds{1}_{y=a}\log(\pi(a \mid o))\bigg]
\end{equation}
where $A$ denotes the action space.

\subsection{Transformers for Decision Making}

\begin{figure}[h]
\begin{center}
\centerline{\includegraphics[width=0.5\textwidth]{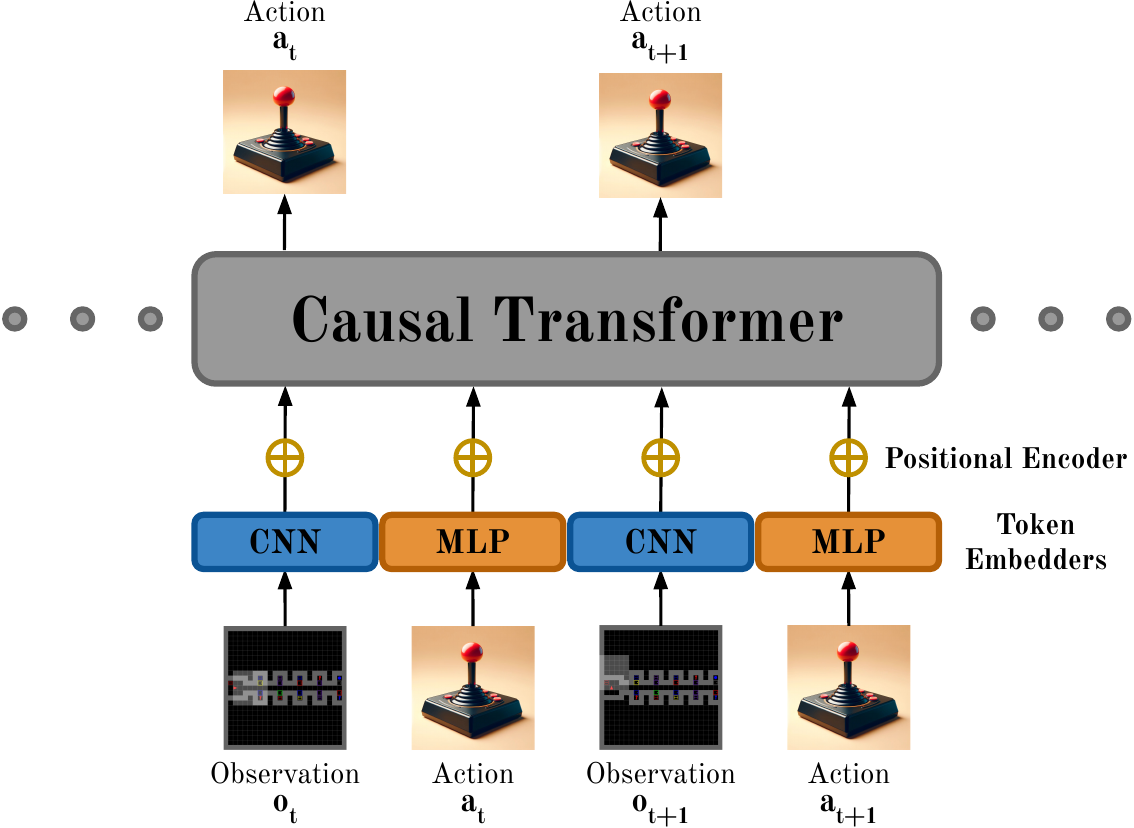}}
\caption{Architecture of the causal Transformer for sequential decision making modeling used in all AttentionTuner and vanilla Transformer experiments. The model embeds observations $o_t$ using a Convolutional Neural Network (CNN) and actions $a_t$ using a Multilayer Perceptron (MLP) network. Positional encodings are added to these embeddings to preserve positional context. Subsequently, they are fed into a causal Transformer decoder, which predicts future actions $a_{t}, a_{t+1}, \ldots$, conditioned on past events. Complete architectural details can be found in Appendix~\ref{architecture}.}
\label{transformer}
\end{center}
\end{figure}

Transformers are a neural network architecture designed to process sequential data \cite{vaswani2023attention}. The core of the Transformer architecture is the self-attention mechanism, which allows the model to weigh the importance of different parts of the input sequence. For any given sequence, the Transformer computes sets of queries $\mathbf{Q}$, keys $\mathbf{K}$, and values $\mathbf{V}$, typically through linear transformations of the input. Attention scores are calculated as follows:
\begin{equation}
    \text{Attention}(\mathbf{Q}, \mathbf{K}, \mathbf{V}) = \text{softmax}\left(\frac{\mathbf{QK}^T}{\sqrt{d_k}}\right)\mathbf{V}
\end{equation}
Here, \(d_k\) represents the dimension of the keys $\mathbf{K}$. To ensure causality in the Transformer model, masking is applied to the attention scores, particularly by zeroing out the upper right triangle of the attention matrix. This process ensures that a token is only influenced by preceding tokens. The self-attention mechanism equips the Transformer with an episodic memory, enabling it to emphasize previous tokens based on the similarity between query and key vectors.

The architecture extends this concept through multi-head self-attention, which involves computing attention scores multiple times with different sets of $\mathbf{Q}$, $\mathbf{K}$, and $\mathbf{V}$. Each Transformer layer consists of this multi-head self-attention mechanism, a feedforward network, and layer normalization. The complete Transformer decoder-only model is formed by stacking multiple such layers. Additionally, to retain positional information of tokens, positional encodings are added to the input embeddings before the first layer.

In the context of decision-making, trajectories can be modeled using the Transformer architecture, as illustrated in Figure~\ref{transformer}. This approach is analogous to the methods used in Decision Transformers \cite{Chen2021DecisionTR} and Trajectory Transformers \cite{Janner2021OfflineRL}. We represent a trajectory as:
\begin{equation}
    \tau = (o_1, a_1, o_2, a_2, \ldots, o_T, a_T)
\end{equation}
where $o_i$ and $a_i$ denote the observation and action, respectively, at timestep \(i\). For token embeddings, observations $o_i$ are processed through a convolutional neural network \cite{lecun1998gradient}, while actions $a_i$ are fed into a linear layer. The Transformer predicts the action $a_t$ for each observation $o_t$ as detailed in Figure~\ref{transformer}. These predictions are then utilized to compute the imitation learning training loss, as described in Equation~\ref{ILloss}.

Although Transformers excel in numerous domains, they encounter specific challenges in POMDP memory tasks. These challenges lead to the difficult and unstable optimization of the behavioral cloning objective (Equation~\ref{ILloss}), an issue further explored in Section~\ref{analysis}.

\section{Method}
\label{method}

We propose a novel approach to imitation learning by introducing \textbf{memory dependency pairs} within trajectories to address memory utilizations in decision-making. For each trajectory $\tau_i \in D$, we define $\mathcal{M}_i = \{(p_j, q_j)\}_{j = 1 \ldots |\mathcal{M}_i|}$ as the set of memory dependency pairs, where each pair $(p, q)$ indicates that observation $o_p$ was recalled during the decision-making process for $a_q$. If the decision-making process for $a_q$ depends on multiple past observations $o_{p_1}, \ldots, o_{p_n}$, then we can represent these dependencies with the pairs $(p_1, q), \ldots, (p_n, q)$. We extend the definition of an expert demonstration trajectory, initially described in Section~\ref{IL_intro}, to $\tau_i = \left(\{(o_j, a_j)\}_{j = 1 \ldots |\tau|}, \mathcal{M}_i\right)$, incorporating both observation-action pairs and memory dependencies. While in practice, a human would typically annotate memory dependency pairs (as shown in Appendix~\ref{sec:annotate}), in the experiments reported in this paper, we instead use a computer program to automate the annotation of memory dependency pairs.

While in principle, memory dependency pairs could be used to enhance any memory-based learning architecture, in this paper we introduce \textbf{AttentionTuner}, which specifically leverages them to enhance Transformer-based architectures. For each trajectory $\tau_i$, with length $n = |\tau_i|$, AttentionTuner constructs an expert self-attention matrix $E \in \{0,1\}^{n \times n}$, detailed in Figure~\ref{fig1}. This matrix, derived from $\mathcal{M}_i$, is defined as:
\[
E_{ij} = 
\begin{cases} 
1 & \text{if } (j, i) \in \mathcal{M}_i \\
0 & \text{otherwise}
\end{cases}
\]
To encourage the Transformer to mimic the expert's memory mechanism, AttentionTuner applies a binary cross-entropy loss between the expert matrix $E$ and the learner's self-attention matrix $A = \sigma(QK^\top) \in [0,1]^{n \times n}$. The memory loss equation is:
\begin{equation}
\label{memloss}
\begin{split}
    \mathcal{L}\left(A, E\right) &= -\frac{1}{n^2}\sum_{i = 1}^n\sum_{j = 1}^n\bigg[
    E_{ij}\log(A_{ij}) \\ &+ (1 - E_{ij})\log(1 - A_{ij})\bigg]
\end{split}
\end{equation}
In AttentionTuner, this memory loss is applied to a single attention head within the first Transformer layer (Figure~\ref{4x2}). The first layer is chosen because it is closest to the raw observation embeddings, making the application of memory dependency pairs more meaningful. Applying the loss to a single head allows other heads to learn additional memory mechanisms not captured by $\mathcal{M}_i$. Alternative applications of this memory loss are explored in Appendix~\ref{sec:mem-ablate}.

The memory loss is then scaled using a hyperparameter $\lambda$ and combined with the imitation learning loss $\mathcal{L_{\mathrm{IL}}}$ (defined in Equation~\ref{ILloss}) to form the final training loss:
\begin{equation}
    \mathcal{L} = \mathcal{L_{\mathrm{IL}}} + \lambda\mathcal{L}(A, E)
\end{equation}
We set $\lambda = 10$ based on robust performance observed across various benchmark tasks, effectively balancing the magnitude of the memory loss $\mathcal{L}(A, E)$ with the imitation learning loss $\mathcal{L_{\mathrm{IL}}}$. Comprehensive details of the model architecture, including the CNN and MLP embedders and the causal Transformer, are provided in Appendix~\ref{architecture}. Pseudocode for training AttentionTuner is provided in Algorithm~\ref{alg:attention_tuner}. The pseudocode for training vanilla Transformers is identical if the memory loss $\mathcal{L}_{\mathrm{memory}}$ is removed by setting $\lambda = 0$.

\begin{algorithm}[h]
\caption{AttentionTuner}
\label{alg:attention_tuner}
\begin{algorithmic}[1]
\REQUIRE Expert demonstrations $D = \{\tau_i\}_{i = 1 \ldots M}$, each $\tau_i = \left(\{(o_j, a_j)\}_{j = 1 \ldots |\tau|}, \mathcal{M}_i\right)$
\REQUIRE Hyperparameter for memory loss scaling $\lambda$
\REQUIRE Learning rate $\eta$
\REQUIRE Transformer model with CNN and MLP embedders, parameterized by $\theta$
\FORALL{epochs}
    \FORALL{$\tau_i \in D$}
        \STATE Extract observation-action pairs and memory dependency pairs: $\{(o_j, a_j)\}$, $\mathcal{M}_i$
        \STATE $o_{\mathrm{emb}} \leftarrow \mathrm{CNN}(\{o_j\})$
        \STATE $a_{\mathrm{emb}} \leftarrow \mathrm{MLP}(\{a_j\})$
        \STATE $\mathrm{input\_seq} \leftarrow \mathrm{PositionalEncoding}(o_{\mathrm{emb}}, a_{\mathrm{emb}})$
        \STATE $\{\hat{a}_j\}, A \leftarrow \mathrm{Transformer}(\mathrm{input\_seq})$ \COMMENT{$A$ is the self-attention matrix $\sigma(QK^\top)$ of the first attention head in the first Transformer layer}
        \STATE Initialize $E \leftarrow \{0\}^{|\tau_i| \times |\tau_i|}$
        \FORALL{$(p, q) \in \mathcal{M}_i$}
            \STATE $E[q][p] = 1$
        \ENDFOR
        \STATE $\mathcal{L}_{\mathrm{memory}} \leftarrow \mathrm{BinaryCrossEntropy}(A, E)$
        \STATE $\mathcal{L}_{\mathrm{IL}} \leftarrow \mathrm{NegativeLogLikelihood}(\{\hat{a}_j\}, \{a_j\})$
        \STATE $\mathcal{L} \leftarrow \mathcal{L}_{\mathrm{IL}} + \lambda \cdot \mathcal{L}_{\mathrm{memory}}$ 
        \STATE $\theta \leftarrow \theta - \eta\nabla_\theta\mathcal{L}$
    \ENDFOR
\ENDFOR
\end{algorithmic}
\end{algorithm}

\section{Experiments}

\begin{figure*}[ht]
\begin{center}
\centerline{\includegraphics[width=\textwidth]{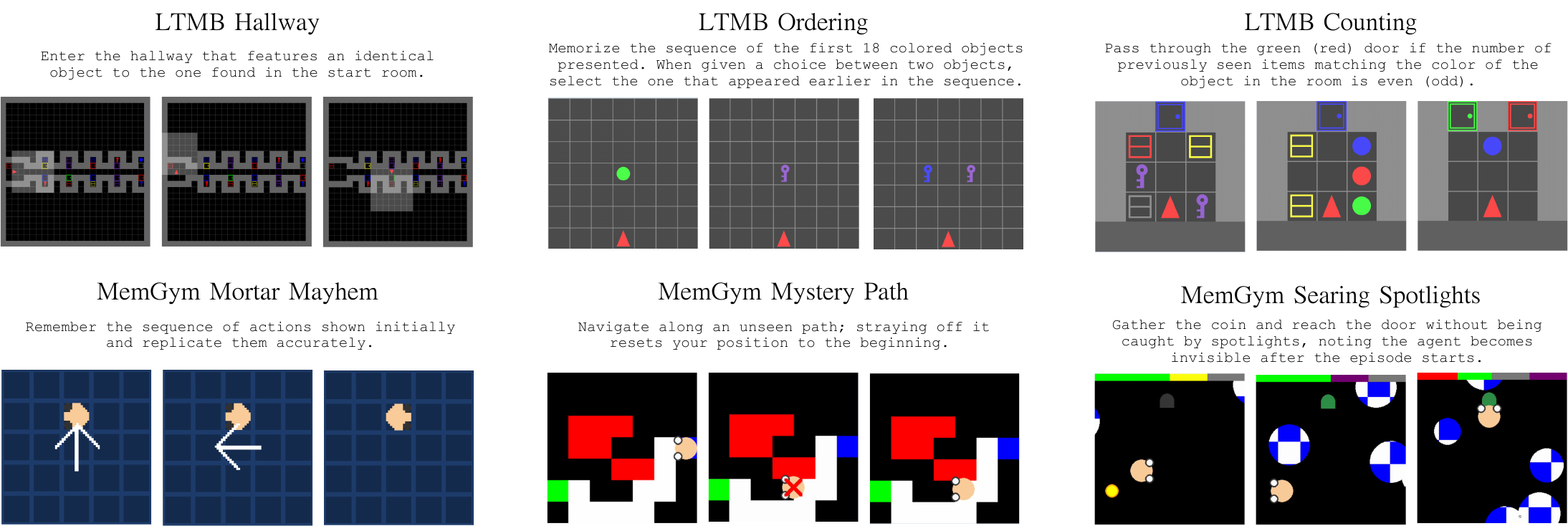}}
\caption{Overview of LTMB and MemGym tasks, each represented by a sequence of three images: the initial observation (left), a subsequent observation that must be recalled (middle), and a decision-making state that depends on memory of the middle state (right). Detailed task descriptions are available in Appendix~\ref{app:descrip}.}
\label{environments}
\end{center}

\end{figure*}

In this section, we conduct empirical assessments to 
\begin{enumerate}
    \item Determine whether AttentionTuner improves success rates and long-term credit assignment compared to the vanilla Transformer.

    \item Evaluate the value of memory dependency pair annotations for training in comparison to additional demonstrations.

    \item Assess the feasibility and human labor costs associated with annotating memory dependency pairs.
\end{enumerate}

\subsection{Experimental Setup}
We evaluated AttentionTuner on the Memory Gym benchmark \cite{pleines2023memory} and our newly introduced Long-term Memory Benchmark (LTMB), each featuring three procedurally generated POMDP tasks as illustrated in Figure~\ref{environments}. These tasks are specifically designed to necessitate and assess the use of memory mechanisms in agents. Four of the six tasks require memory dependencies on several past observations at a single timestep (shown in Appendix \ref{expert} Figure \ref{expert_heatmaps}). A key criterion for selecting these benchmarks was their simple environment dynamics, allowing a focus on episodic memory aspects within the tasks. Preference was given to gridworld environments whenever possible to simplify the process of hand-designing expert policies and automating memory dependency pair annotations. 

\paragraph{Baselines} Due to the lack of a good dense reward function in our benchmark environments and in most real world decision making tasks, we do not directly compare our method against methods that learn memory mechanisms through reinforcement learning, as they are expected to perform poorly on these tasks. We believe AttentionTuner focuses on a new problem setting, and we are not aware of other methods for learning memory mechanisms through demonstrations. For this reason, we compare our method against a vanilla Transformer as a baseline.

All experiments were conducted using at least 5 random seeds (Appendix~\ref{rand}), with each evaluated on 1,000 trials. Illustrations of sample expert attention matrices $E$ for all tasks are provided in Appendix~\ref{expert} Figure~\ref{expert_heatmaps}. Detailed environment settings and data collection for each task are provided in Appendix~\ref{settings}. Additionally, Appendix~\ref{hyperparameters} comprehensively documents the training hyperparameters utilized in our experiments. Finally, Appendix~\ref{compute} lists the computational resources allocated for these experiments.

\subsection{Results and Analysis}
\label{analysis}

\begin{table*}[ht]
\begin{center}
\resizebox{\textwidth}{!}{
\begin{small}
\begin{sc}
\begin{tabular}{lccccccr}
\toprule
    & Mortar & Mystery & Searing & & & \\
Methods & Mayhem & Path & Spotlights & Hallway & Ordering & Counting \\
\midrule
Vanilla Transformer & $20.8 \pm 42.2$ & $97.3 \pm 0.7$ & $62.2 \pm 5.5$ & $53.2 \pm 28.3$ & $59.4 \pm 22.9$ & $6 \pm 0.7$ \\
AttentionTuner (ours) & $\textbf{99.8} \pm 0.4$ & $\textbf{98.7} \pm 0.4$ & $\textbf{64.2} \pm 3.4$ & $\textbf{99.9} \pm 0.1$ & $\textbf{99.9} \pm 0.3$ & $\textbf{6.5} \pm 0.4$ \\
\bottomrule
\end{tabular}
\end{sc}
\end{small}
}
\end{center}
\caption{Average success rates and 90\% confidence intervals for two different methods—Vanilla Transformer and AttentionTuner (our approach)—across various tasks in the Memory Gym and Long-term Memory Benchmark.}
\label{benchmark}
\end{table*}

\begin{figure*}[ht]
\begin{center}
\centerline{\includegraphics[width=\textwidth]{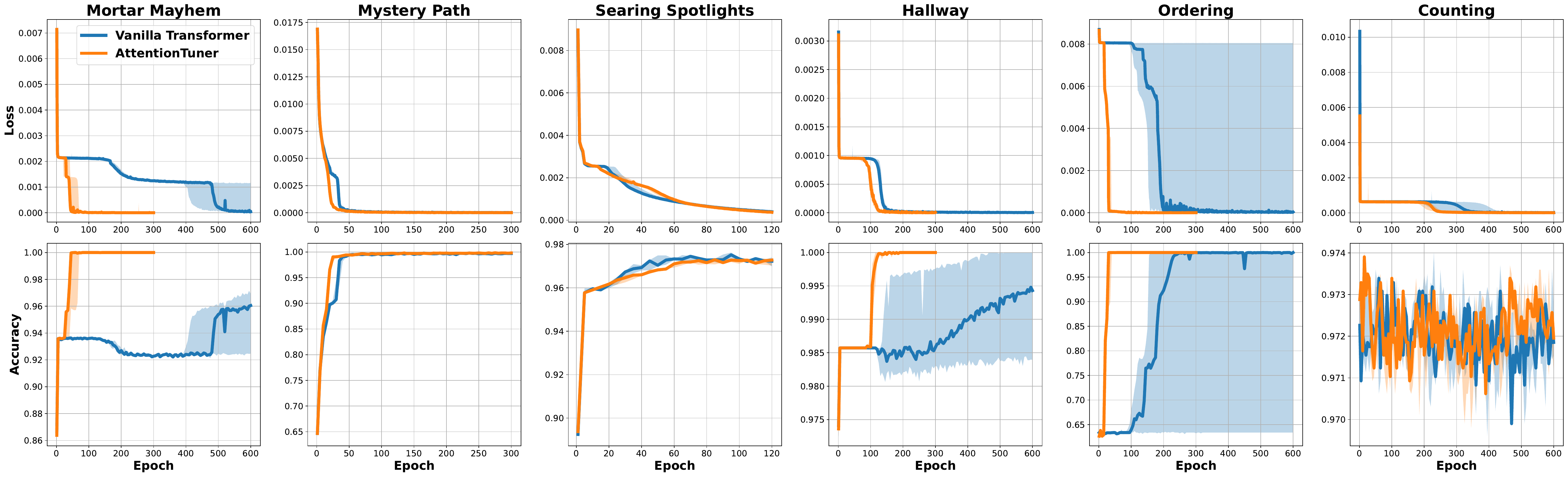}}
\caption{Median learning curves with interquartile range for Memory Gym and LTMB tasks are presented. The median is used rather than the mean due to the influence of outliers. Detailed mean learning curves are accessible in Appendix~\ref{mean_curves_section} Figure~\ref{mean_curves}. The top row displays the training loss while the bottom row shows the test accuracy on action prediction (not success rate). For a direct comparison with the vanilla Transformer, only the imitation learning loss (Equation~\ref{ILloss}) is plotted for AttentionTuner.} 
\label{med_curves}
\end{center}
\end{figure*}

\paragraph{AttentionTuner achieves significantly better success rates.}
In Table~\ref{benchmark}, performance of AttentionTuner is compared with the vanilla Transformer on Memory Gym and LTMB tasks. We performed a Welch t-test (results in Appendix~\ref{welch} Table~\ref{ttest}) with a significance threshold of $\alpha = 0.05$ and found that AttentionTuner achieves significantly better success rates on Mortar Mayhem, Mystery Path, Hallway, and Ordering. The high variability observed in the vanilla Transformer's performance on Mortar Mayhem, Hallway, and Ordering tasks can be attributed to some runs achieving nearly perfect success rates, while others fall close to zero.  This discrepancy is explained in the learning curves presented in Figure~\ref{med_curves}, indicating challenges in optimizing the behavioral cloning objective (Equation~\ref{ILloss}) for POMDP memory tasks with a large Transformer model. These challenges result in convergence to suboptimal local optima, as evidenced by flat segments in the training loss curves. 

\paragraph{AttentionTuner aids the optimizer in navigating the loss landscape.}
A local optimum is apparent in the Hallway and Counting tasks, while less pronounced in Mystery Path and Searing Spotlights (Figure~\ref{med_curves}). Notably, Mortar Mayhem and Ordering present two local optima, posing additional challenges for the optimizer. A unique observation in the Ordering task is that AttentionTuner encounters a single local optimum, in contrast to the vanilla Transformer, which faces two. The intricacies of task design that lead to the formation of multiple or difficult-to-escape local optima are not fully understood, highlighting an area for future research. 

The interquartile ranges in Figure~\ref{med_curves} suggest that AttentionTuner aids the optimizer in more efficiently and consistently escaping local optima, and potentially encountering fewer of them. For instance, in the Mortar Mayhem task, AttentionTuner enabled the optimizer to surpass both local optima in under 50 epochs with minimal variability across training seeds. In contrast, only a single training run of the vanilla Transformer overcame both local optima and did so with $\sim 400$ more training epochs. A similar pattern emerges in the Mystery Path, Hallway, Ordering, and Counting tasks where AttentionTuner consistently escapes from the local optima while the vanilla Transformer only escapes some of the time or escapes up to 200 epochs later than AttentionTuner (Figure~\ref{med_curves}). AttentionTuner's enhanced capability in traversing the loss landscape underscores its efficacy in facilitating long-term credit assignment and the learning of memory mechanisms.

\paragraph{AttentionTuner promotes convergence to solutions with better generalizabiltiy.}
Despite nearly identical learning curves after epoch 50 (Figure~\ref{med_curves}) on Mystery Path, AttentionTuner registers a significantly higher success rate than the vanilla Transformer. This difference in success rate suggests that even when the memory loss does not significantly improve training, it still promotes convergence to solutions with better generalizability. The differences in the final memory mechanisms learned by each model are illustrated in Appendix~\ref{learned_heatmaps}.

\paragraph{AttentionTuner exhibits limitations in learning short-term memory mechanisms.} In the Searing Spotlights task, AttentionTuner does not show a significant improvement in success rate or training efficiency. The expert attention matrix $E$ for this task (shown in Appendix~\ref{expert}), which attends to all previous actions is more resemblent of short-term memory than sparser long-term memory that our method is designed to optimize. Training an attention head to focus on all previous tokens effectively amounts to not focusing on any specific token, thereby diminishing the effectiveness of the memory loss. Despite efforts to distribute the memory loss across various attention heads, as investigated in Appendix \ref{sec:mem-ablate} Table~\ref{mem_ablation}, we did not observe any significant improvements. This outcome reinforces our understanding that the memory dynamics in the Searing Spotlights task are not ideally suited to the capabilities of AttentionTuner.

\subsection{Ablations}
\label{ablation}

In this section, we use ablations to assess the practicality of annotating memory dependency pairs. For each ablation study, experiments were run on the Mortar Mayhem and Hallway tasks with 5 random seeds.

\subsubsection{Imperfect Expert Annotations.}
\label{sec:annotate_ablate}

\begin{figure}[ht]
\begin{center}
\centerline{\includegraphics[width=0.5\textwidth, keepaspectratio]{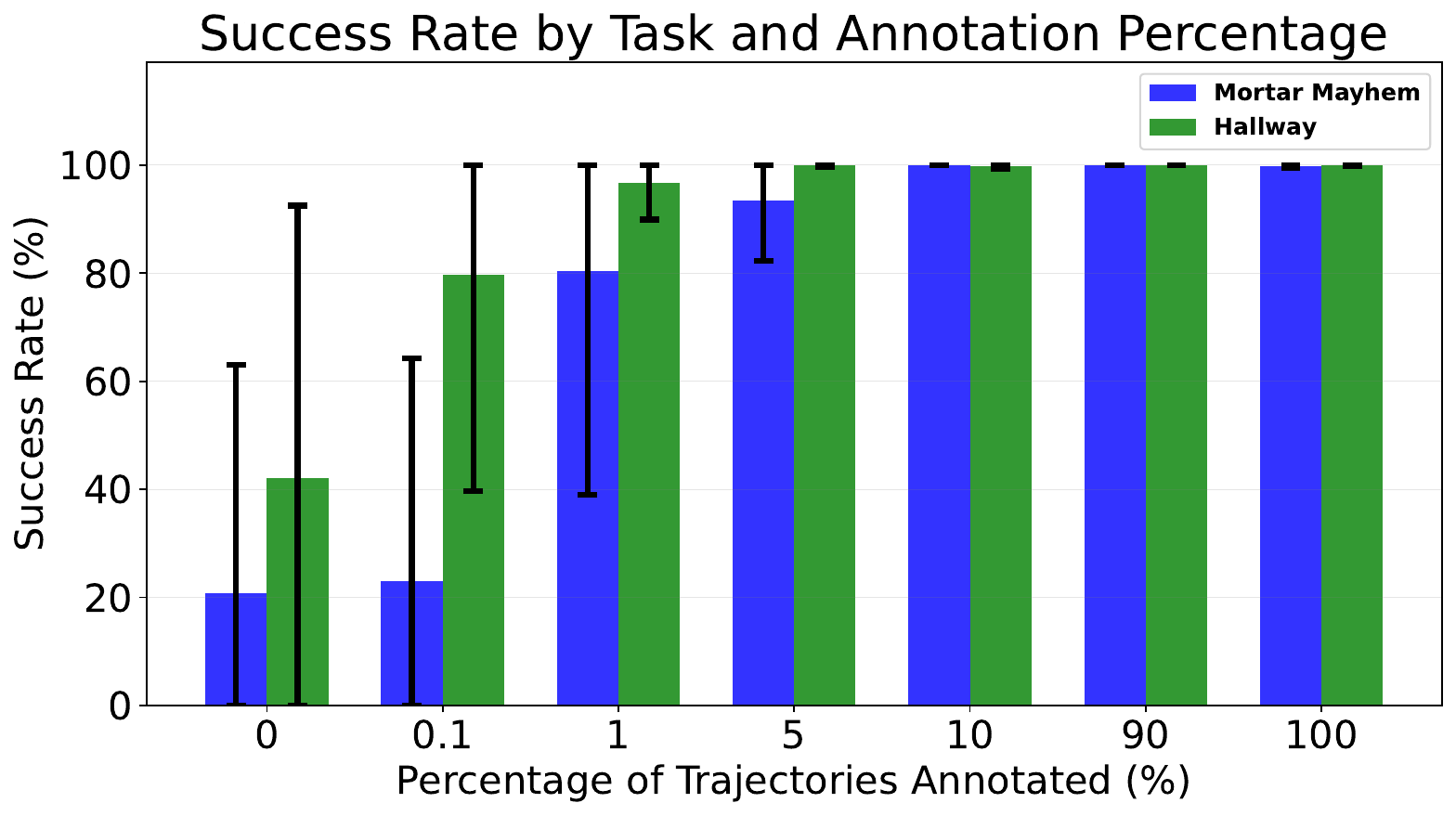}}
\caption{Success rates and 90\% confidence intervals for AttentionTuner training on 'Mortar Mayhem' and 'Hallway' tasks with missing annotations. Numerical values are presented in Appendix~\ref{sec:annotation} Table~\ref{tbl:annotations_ablation}.}
\label{annotations_ablation}
\end{center}

\end{figure}

Collecting comprehensive expert memory annotations for every demonstration trajectory in long complex decision-making tasks may not always be feasible. To explore the impact of this constraint, we conducted experiments varying the proportion of trajectories that included memory annotations, as detailed in Figure~\ref{annotations_ablation}. During training, the memory loss (Equation~\ref{memloss}) was only used on trajectories that included memory annotations. Trajectories without memory annotations were trained using only the imitation learning loss (Equation~\ref{ILloss}). The findings, presented in Figure~\ref{annotations_ablation}, reveal that having memory annotations for merely 10\% of the trajectories achieves comparable results to having annotations for all trajectories. Performance degradation for AttentionTuner becomes noticeable when the proportion of annotated demonstration trajectories falls below 10\%. Nonetheless, even with as few as 1\% of trajectories annotated, AttentionTuner manages to achieve a relatively high average success rate. Moreover, having even just 0.1\% of trajectories annotated still enables AttentionTuner to outperform the vanilla Transformer. 

Furthermore, human annotated trajectories could include errors. We found that AttentionTuner is robust against small perturbations of the memory dependency pair endpoints (Appendix~\ref{sec:imprecise}). Potential solutions to make AttentionTuner more robust against larger perturbations are discussed in Section~\ref{sec:summary}.

\subsubsection{Value of Memory Dependency Pairs Compared to Additional Demonstrations.}
\label{sec:value}

\begin{figure}[ht]
    \centering
    \includegraphics[width=0.5\textwidth, keepaspectratio]{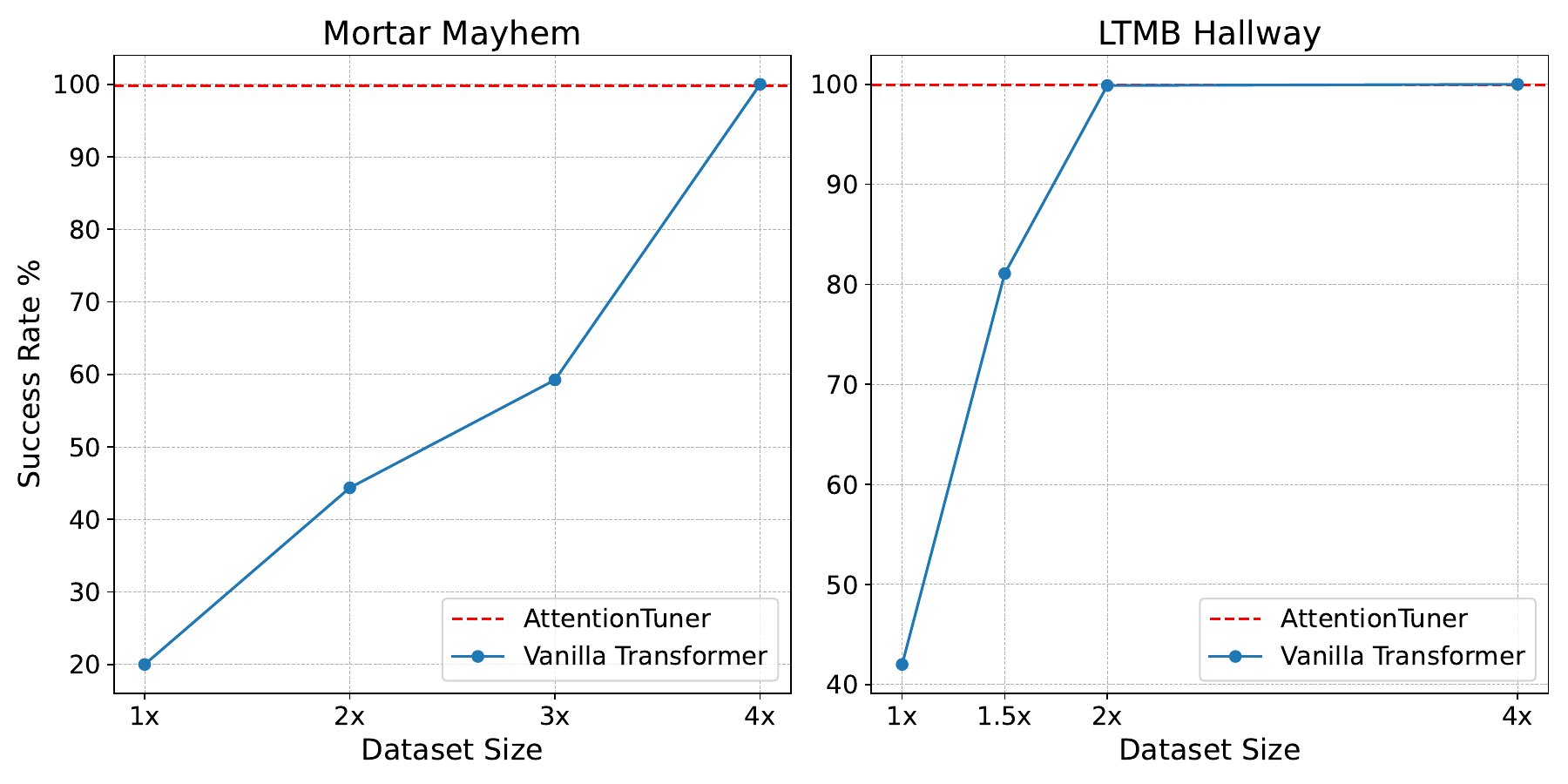}
    \caption{Success rates for vanilla Transformer with different training data sizes. The dotted red line represents AttentionTuner's success rate when training with a 1x dataset size. The number of demonstrations collected in the 1x datasets can be found in Appendix~\ref{settings}.}
    \label{fig:scaling}
\end{figure}

Figure~\ref{fig:scaling} shows that it can take anywhere from 2 to 4 times more demonstrations to train a vanilla Transformer to achieve the same success rate as AttentionTuner. In our experiments, we found that it takes 2 to 3 times longer for a human to annotate memory dependency pairs than to just collect a demonstration (Appendix~\ref{sec:annotate}). Considering that only 5-10\% of the demonstrations need to be annotated (Section~\ref{sec:annotate_ablate}), we find that memory dependency pair annotations on a single demonstration is equivalent to providing 40 additional demonstrations on both the Mortar Mayhem and Hallway tasks. Annotating memory dependency pairs instead of collecting additional demonstrations results in a human labor time savings factor of 16 on Mortar Mayhem and 14 on Hallway. Appendix~\ref{sec:math} details how these numbers are computed, though it is important to note that exact time savings can vary widely based on the task and the quality of the human annotator.

\section{Related Work}
\label{related}

This section surveys the literature on memory types, integration of human feedback in decision-making algorithms, challenges in long-term credit assignment, and the development of memory mechanisms in learning models.

\subsection{Types of Memory}

Endel Tulving's 1972 research distinguishes between episodic memory, which stores information about specific events and their context, and semantic memory, a structured knowledge base for language and symbols \cite{Tulving1972, Tulving1983, tulving1985memory}. Long-term memory preserves a wide array of information, from skills to life events, over lengthy durations. In contrast, short-term or working memory, crucial for tasks like language comprehension and problem-solving, holds information briefly and allows for its manipulation, such as in mental arithmetic \cite{baddeley1992working, cowan2008differences}. While other dichotomies exist in the study of memory, such as declarative versus procedural \cite{humphreys1989different, ten1999procedural, ullman2004contributions}, and active versus inactive \cite{lewis1979psychobiology}, this work is primarily concerned with long-term episodic memory.

\subsection{Human Feedback in Decision Making}

Human feedback can be integrated into learning agents through various modalities. In reinforcement learning, scalar rewards may be assigned to individual states \cite{KCAP09-knox, SuttonBarto2018, warnell2018deep} or preferences may be expressed between trajectory pairs, either online or offline \cite{Wilson2012ABA, akrour2012april, wirth2016model, Sadigh2017ActivePL, lee2021pebble, stiennon2022learning, christiano2023deep}. In imitation learning, agents can learn from observation-action pairs, provided by humans in both online and offline contexts \cite{ross2011reduction, Zhang2017QueryEfficientIL, saunders2017trial, Torabi2018BehavioralCF}. Additional feedback mechanisms include gaze tracking \cite{ 8967843, saran2020efficiently}, binary corrective signals \cite{Celemin2019AnIF}, and human-provided trajectory outlines \cite{gu2023rttrajectory}. To the best of our knowledge, memory dependency pairs are the first modality through which humans can articulate their memory processes for decision-making.

\subsection{Long-Term Credit Assignment}
\label{credit}

The Long-Term Credit Assignment Problem highlights the difficulty of training agents to attribute consequences of actions over extended time frames \cite{sutton1984temporal, bengio1993credit, bengio1994learning}. Humans can base decisions on events from years or even decades past. However, agents today struggle with credit assignment over even short horizons. This challenge primarily arises from vanishing or exploding gradients during backpropagation through time or the memory mechanism's inability to retain relevant information amidst noise \cite{bengio1993credit}. Proposed solutions include adding skip connections to reduce backpropagation distances \cite{ke2018sparse, hung2019optimizing} and using self-attention in transformers \cite{vaswani2023attention}, which allows direct gradient flow between relevant timesteps. However, self-attention has been shown to not improve long-term credit assignment nor fully exploit all available information in its context \cite{liu2023lost, ni2023transformers}. In this work, memory dependency pairs are shown to assist self-attention in long-term credit assignment.

\subsection{Learning Memory Mechanisms}

RNNs were initially augmented with memory by incorporating hidden states and gating mechanisms, such as in Long Short-Term Memory (LSTM) networks \cite{Hochreiter1997LongSM, cho2014learning, burtsev2021memory}. Other approaches include integrating RNNs with differentiable memory that is key-addressable \cite{graves2014neural, weston2015memory, graves2016hybrid, wayne2018unsupervised}. Some researchers have also experimented with augmenting RNNs with stack-like memory modules \cite{joulin2015inferring}. Furthermore, combining LSTMs for short-term working memory with key-addressable memory for long-term episodic memory has been explored \cite{fortunato2020generalization}. Another significant development is the integration of Transformers with differentiable memory that can be either key or content addressable \cite{kang2023think, bessonov2023recurrent}. \citet{allen2024mitigating} uses the difference in temporal difference and Monte Carlo value estimates to detect partial observability and improve memory retention for RNNs. Our work is the first to explore learning memory mechanisms through expert demonstrations.

\section{Summary and Future Work}
\label{sec:summary}

This study introduces \textbf{memory dependency pairs} to enhance long-term credit assignment in POMDP tasks necessitating complex memory mechanisms. We proposed AttentionTuner, a method designed for self-attention architectures to effectively leverage memory dependency pairs. Empirical evaluation on Memory Gym and LTMB benchmarks demonstrate that AttentionTuner effectively mitigates the local optima challenges in memory tasks, either by accelerating escape from such optima or by circumventing them entirely (Figure~\ref{med_curves}). This optimization improvement significantly increases success metrics across various tasks compared to vanilla Transformers (Table~\ref{benchmark}). Notably, these learning benefits are achievable with minimal annotation, requiring as few as $0.1\%$ of training trajectories to be annotated. This level of efficiency makes AttentionTuner a practical tool for real-world applications where learning complex memory mechanisms poses significant challenges.

Our approach primarily aims to enhance long-term episodic memory, thriving particularly when the expert attention matrix $E$ exhibits sparsity. However, it encounters limitations in scenarios involving short-term or semantic memory, a challenge exemplified by the performance in the Searing Spotlights task (Table~\ref{benchmark}). To mitigate this limitation, incorporating a short-term memory mechanism, like frame stacking, could be an effective strategy to complement AttentionTuner's long-term episodic memory.

The practicality of pinpointing precise timesteps for memory dependency pairs becomes cumbersome with longer episode horizons. To address this issue, a plausible direction could involve the summarization of token sequences into more generalized and abstract representations such as done in HCAM \cite{lampinen2021towards}. These summary ``token chunks" would allow for annotators to connect two events with natural language or approximate timesteps instead of having to connect two exact timesteps.

Another constraint stems from the Transformer model's mechanism of incorporating all preceding observations into its context, posing scalability challenges for tasks with extended horizons. Exploring hierarchical attention mechanisms or adopting a key-value cache system \cite{wu2022memorizing} presents promising avenues. Memory dependency pairs could serve as valuable assets in these contexts, guiding the prioritization and retention of pivotal events within each hierarchical layer or assisting in the optimization of key-value cache retrieval and management strategies.

While the focus of this work has been on integrating memory dependency pairs within the Transformer architecture, memory dependency pairs are applicable to a variety of neural architectures. For instance, in RNNs, a reconstruction loss on hidden states could promote memory retention, while in key-addressable, differentiable memory systems, a loss could encourage accurate key additions and queries. State space models can be viewed as minimizing an online learning objective \cite{liu2024longhornstatespacemodels}, and therefore memory dependency pairs can be used to emphasize which tokens the model should prioritize for retention (like a re-weighted regret). These ideas are left as an exciting frontier for future research endeavors.

\section*{Acknowledgements}

This work has taken place in the Learning Agents Research
Group (LARG) at the Artificial Intelligence Laboratory, The University
of Texas at Austin.  LARG research is supported in part by the
National Science Foundation (FAIN-2019844, NRT-2125858), the Office of
Naval Research (N00014-18-2243), Army Research Office (E2061621),
Bosch, Lockheed Martin, and Good Systems, a research grand challenge
at the University of Texas at Austin.  The views and conclusions
contained in this document are those of the authors alone.  Peter
Stone serves as the Executive Director of Sony AI America and receives
financial compensation for this work.  The terms of this arrangement
have been reviewed and approved by the University of Texas at Austin
in accordance with its policy on objectivity in research.

\newpage
\appendix
\onecolumn
\section{Neural Network Architectures}
\label{architecture}
Only the observation and action embedders differ in architecture between the two benchmarks. For both benchmarks, 4 Transformer layers were used with 2 self-attention heads per layer. Additionally, $d_{\text{model}} = 512$ for all experiments.  

\subsection{Causal Transformer}
\begin{figure}[H]
\begin{center}
\centerline{\includegraphics[width=\textwidth, height=0.8\textheight, keepaspectratio]{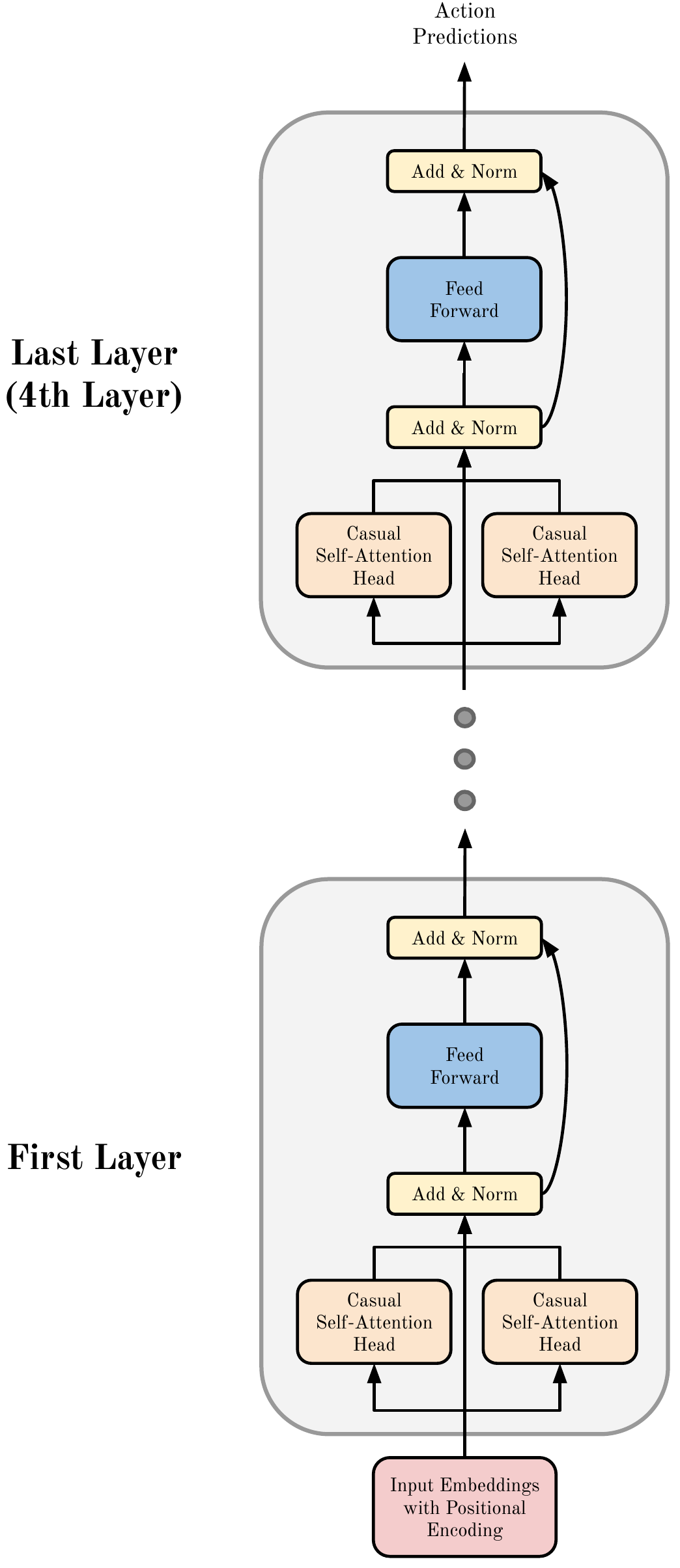}}
\caption{Causal Transformer architecture with 4 layers and 2 self-attention heads per layer as used in all experiments. The memory loss (Equation~\ref{memloss}) was applied to a single self-attention head in the first Transformer layer.}
\label{4x2}
\end{center}

\end{figure}

\subsection{Memory Gym}
\begin{lstlisting}
Transformer(
  (image_embedding): ImageEmbedding(
    (cnn): Sequential(
      (0): Conv2d(3, 32, kernel_size=(8, 8), stride=(4, 4))
      (1): ReLU()
      (2): Conv2d(32, 64, kernel_size=(4, 4), stride=(2, 2))
      (3): ReLU()
      (4): Conv2d(64, 64, kernel_size=(3, 3), stride=(1, 1))
      (5): ReLU()
    )
    (fc): Sequential(
      (0): Linear(in_features=3136, out_features=512, bias=True)
      (1): Tanh()
    )
  )
  (action_embedding): ActionEmbedding(
    (mlp): Sequential(
      (0): Linear(in_features=4, out_features=512, bias=True)
      (1): Tanh()
    )
  )
  (positional_encoding): PositionalEmbedding()
  (embedding_LN): LayerNorm((512,), eps=1e-05, elementwise_affine=True)
  (dropout): Dropout(p=0.1, inplace=False)
  (transformer_layers): ModuleList(
    (0-3): 4 x TransformerLayer(
      (self_attention): MultiheadAttention(
        (out_proj): NonDynamicallyQuantizableLinear(in_features=512, out_features=512, bias=True)
      )
      (norm1): LayerNorm((512,), eps=1e-05, elementwise_affine=True)
      (feedforward): Sequential(
        (0): Linear(in_features=512, out_features=2048, bias=True)
        (1): ReLU()
        (2): Linear(in_features=2048, out_features=512, bias=True)
        (3): Dropout(p=0.1, inplace=False)
      )
      (norm2): LayerNorm((512,), eps=1e-05, elementwise_affine=True)
    )
  )
  (output): Linear(in_features=512, out_features=4, bias=True)
)
\end{lstlisting}

\subsection{LTMB}
\begin{lstlisting}
Transformer(
  (image_embedding): ImageEmbedding(
    (cnn): Sequential(
      (0): Conv2d(20, 40, kernel_size=(3, 3), stride=(1, 1), padding=(1, 1))
      (1): ReLU()
      (2): Conv2d(40, 80, kernel_size=(3, 3), stride=(1, 1), padding=(1, 1))
      (3): ReLU()
    )
    (fc): Sequential(
      (0): Linear(in_features=3920, out_features=512, bias=True)
      (1): Tanh()
    )
  )
  (action_embedding): ActionEmbedding(
    (mlp): Sequential(
      (0): Linear(in_features=7, out_features=512, bias=True)
      (1): Tanh()
    )
  )
  (positional_encoding): PositionalEmbedding()
  (embedding_LN): LayerNorm((512,), eps=1e-05, elementwise_affine=True)
  (dropout): Dropout(p=0.1, inplace=False)
  (transformer_layers): ModuleList(
    (0-3): 4 x TransformerLayer(
      (self_attention): MultiheadAttention(
        (out_proj): NonDynamicallyQuantizableLinear(in_features=512, out_features=512, bias=True)
      )
      (norm1): LayerNorm((512,), eps=1e-05, elementwise_affine=True)
      (feedforward): Sequential(
        (0): Linear(in_features=512, out_features=2048, bias=True)
        (1): ReLU()
        (2): Linear(in_features=2048, out_features=512, bias=True)
        (3): Dropout(p=0.1, inplace=False)
      )
      (norm2): LayerNorm((512,), eps=1e-05, elementwise_affine=True)
    )
  )
  (output): Linear(in_features=512, out_features=7, bias=True)
)
\end{lstlisting}

\section{Application of the Memory Loss in Transformers}
\label{sec:mem-ablate}

\begin{table*}[ht]
\begin{center}
\resizebox{\textwidth}{!}{
\begin{small}
\begin{sc}
\begin{tabular}{lcccccccr}
\toprule
    & No Layer & First Layer & Middle Layer & Last Layer & First Layer & Middle Layer & Last Layer \\
Tasks & No Heads & All Heads & All Heads & All Heads & Single Heads & Single Heads & Single Heads \\
\midrule
Mortar Mayhem & $20.8 \pm 42.2$ & $\textbf{100} \pm 0.1$ & $99.7 \pm 0.6$ & $99.9 \pm 0.1$ & $99.8 \pm 0.4$ & $100 \pm 0.1$ & $39.9 \pm 52.1$ \\
Hallway & $42 \pm 50.5$ & $99.9 \pm 0.2$ & $\textbf{100} \pm 0$ & $89.4 \pm 22.5$ & $99.9 \pm 0.1$ & $99.8 \pm 0.3$ & $99.12 \pm 1.7$ \\
\bottomrule
\end{tabular}
\end{sc}
\end{small}
}
\end{center}
\caption{Success rate and 90\% confidence interval achieved on memory loss ablations}
\label{mem_ablation}
\end{table*}

The memory loss in Equation~\ref{memloss} has to be applied to a self-attention head in the Transformer. We posited in Section~\ref{method} that applying the memory loss to the first Transformer layer would make the memory dependency pairs more meaningful as the self-attention mechanism attends over the raw observation embeddings. This hypothesis is supported by the results in Table~\ref{mem_ablation}, which demonstrate that implementations applying the memory loss to the first layer consistently yield near-perfect success rates, surpassing other configurations. While we also speculated that dedicating memory loss to a single attention head would permit the remaining heads to engage in other memory processes, this distinction was not markedly evident in our results. The probable explanation is that memory dependency pairs in these relatively simple tasks sufficiently encapsulate all necessary memory functions, diminishing the benefit of isolating the memory loss to a single head. Attempting to distribute memory dependency pairs among various heads (Appendix~\ref{expert} Figure~\ref{split}), especially in the context of Searing Spotlights with its large amount of memory dependency pairs, did not yield a notable improvement in performance.

\section{Environment Descriptions}
\label{app:descrip}

\subsection{Memory Gym}
Memory Gym features three tasks—Mortar Mayhem, Mystery Path, and Searing Spotlights—set within a $7 \times 7$ gridworld. Agents receive $84 \times 84$ RGB image observations of the gridworld. For Mortar Mayhem and Mystery Path, the discrete action space includes: \texttt{move forward}, \texttt{turn left}, \texttt{turn right}, and \texttt{nop}. Searing Spotlights employs a multi-discrete action space, allowing movement in cardinal or ordinal directions plus a \texttt{nop} option.

\paragraph{Mortar Mayhem}
In this task, the agent memorizes and later executes a sequence of commands, indicated by arrows. An expert would annotate memory dependency pairs $(p, q)$, with $o_p$ representing the observation displaying the command and $o_q$ the observation of its execution.

\paragraph{Mystery Path}
Agents navigate a gridworld with an invisible path, restarting from the beginning if they deviate. To progress, they must remember their path to the deviation point. An expert would annotate memory dependency pairs $(p, q)$ where $o_p$ is a cell adjacent to the cell at $o_q$.

\paragraph{Searing Spotlights}
Agents aim to reach a door in a 2D plane after collecting a key, initially visible but obscured after 6 timesteps by dimming lights. The agent has to make it to the key and door while avoiding ``searing spotlights". An expert would annotate memory dependency pairs $(0, q), (1, q), \ldots, (q - 1, q)$ as agents must recall their starting position and all previous actions to deduce their current location. 

\subsection{Long-term Memory Benchmark}
The Long-term Memory Benchmark (LTMB) comprises three tasks: Hallway, Ordering, and Counting, set in the Minigrid environment \cite{MinigridMiniworld23}. Agents navigate a gridworld, receiving standard Minigrid $7 \times 7 \times 3$ state-based observations. The discrete action space includes \texttt{move forward}, \texttt{turn left}, \texttt{turn right}, and \texttt{nop}.

\paragraph{Hallway}
Agents identify and enter a hallway with an object matching one in the start room. The agent's view is limited to the $7 \times 7$ grid ahead. The expert annotates $(1, q)$ where timestep 1 represents the initial object observation.

\paragraph{Ordering}
Agents memorize the sequence of the first 18 colored objects encountered. When choosing between two objects, the agent selects the one appearing earlier in the sequence. An expert would annotate memory dependency pairs $(p, q)$, connecting the first observation $o_p$ of an object to the query observation $o_q$.

\paragraph{Counting}
In this task, agents traverse gallery rooms, memorizing six objects each, and query rooms, deciding which door to pass based on the parity of a query object's previous appearances. Passing through the wrong door ends the episode. An expert would annotate $(p, q)$ where $o_p$ is a gallery room containing the query object and $o_q$ is the query room.

\section{Environment Settings}
\label{settings}

Memory dependency pairs are annotated for every expert trajectory collected. In practice, memory dependency pairs would need to be annotated by the human demonstrator, for example via a graphical user interface of some sort. While it remains to be verified that this process can be made relatively seemless for human experts, for the purposes of this paper, we sidestep this human-computer interaction issue by simulating both the expert demonstrations and the annotation of memory dependency pairs.

\subsection{Mortar Mayhem}
Discrete action movements were used with a command count of 10. Settings were default, except where noted. A total of 4,000 expert trajectories were collected, each with 118 timesteps.

\subsection{Mystery Path}
The origin and goal were not shown, with other settings at default. Training involved 4,000 expert trajectories, averaging 43 timesteps, and reaching up to 128 timesteps.

\subsection{Searing Spotlights}
The agent was visible for 6 timesteps before the lights dimmed completely. A single coin was used to unlock the exit. Other settings were left at the default. A single coin unlocked the exit, and other settings remained default. Training included 40,000 expert trajectories, averaging 30 timesteps, with a maximum of 75.

\subsection{Hallway}
The environment length was set to 30. A total of 5,000 expert trajectories were collected, averaging 67 timesteps, and maxing at 145 timesteps.

\subsection{Ordering}
The environment length was set to 50. A total of 5,000 expert trajectories were collected, each with a length of 68 timesteps.

\subsection{Counting}
The environment length was 20, with test room frequency at 30\% and empty tile frequency at 10\%. A total of 10,000 expert trajectories were collected, averaging 97 timesteps, with a maximum of 140.

\section{Training Hyperparameters}
\label{hyperparameters}

The following hyperparameters were shared by both AttentionTuner and the vanilla Transformer across all experiments:
\begin{table}[H]
\centering
\begin{tabular}{lll}
\toprule
\textbf{Hyperparameter} & \textbf{Value} & \textbf{Brief Description} \\
\midrule
batch size              & 64             & number of samples in each training iteration \\
learning rate           & $10^{-4}$      & learning rate for gradient descent \\
optimization algorithm  & Adam           & optimization algorithm used \\
$\beta_1$               & 0.9            & exponential decay rate for first moment estimates in Adam \\
$\beta_2$               & 0.999          & exponential decay rate for second moment estimates in Adam \\
epsilon                 & $10^{-8}$      & small constant for numerical stability \\
weight decay            & 0              & weight regularization \\
$\lambda$               & 10             & memory loss multiplier defined in Section~\ref{method} (only for AttentionTuner) \\
\bottomrule
\end{tabular}
\caption{Hyperparameters used in experiments along with their brief descriptions}
\label{params}
\end{table}

Only the number of training epochs differed between methods and tasks.
\begin{table}[H]
\centering
\begin{tabular}{lcc}
\toprule
\textbf{Task}           & \textbf{AttentionTuner} & \textbf{Vanilla Transformer} \\
\midrule
Mortar Mayhem           & 300                     & 600                           \\
Mystery Path            & 300                     & 300                           \\
Searing Spotlights      & 120                     & 120                           \\
Hallway                 & 300                     & 600                           \\
Ordering                & 300                     & 600                           \\
Counting                & 600                     & 600                           \\
\bottomrule
\end{tabular}
\caption{Number of training epochs used for each task}
\label{epochs}
\end{table}

All experiments in Table~\ref{mem_ablation} (Section~\ref{sec:mem-ablate}) used 300 training epochs. All experiments in Figure~\ref{annotations_ablation} (Section~\ref{sec:annotate_ablate}) used 300 training epochs expect for vanilla Transformer runs (0\%) and Mortar Mayhem 0.1\% and 1\% which used 600 epochs.

\section{Random Seeds}
\label{rand}

All experiments were run with random seeds 1 through 5 except for the following:
\begin{itemize}
    \item 10 random seeds were used for vanilla Transformer on LTMB's Hallway task in Table~\ref{benchmark}.
    \item 15 random seeds were used for vanilla Transformer on LTMB's Ordering task in Table~\ref{benchmark}.
\end{itemize}
More training runs were used for these experiments due to their high variability.

\section{Welch t-test for Statistical Significance}
\label{welch}

We applied the Welch t-test to assess the statistical significance of the performance differences between AttentionTuner and the vanilla Transformer, as reported in Table~\ref{benchmark}.

\begin{table}[H]
\centering
\begin{tabular}{lc}
\toprule
\textbf{Task}           & \textbf{p-value} \\
\midrule
Mortar Mayhem           & 0.016 \\
Mystery Path            & 0.012 \\
Searing Spotlights      & 0.546 \\
Hallway                 & 0.014 \\
Ordering                & 0.008 \\
Counting                & 0.261 \\
\bottomrule
\end{tabular}
\caption{Welch t-test p-values for Performance Comparison Across Tasks}
\label{ttest}
\end{table}

\section{Average Learning Curves with 90\% Confidence Intervals}
\label{mean_curves_section}

\begin{figure}[H]
\begin{center}
\centerline{\includegraphics[width=\textwidth]{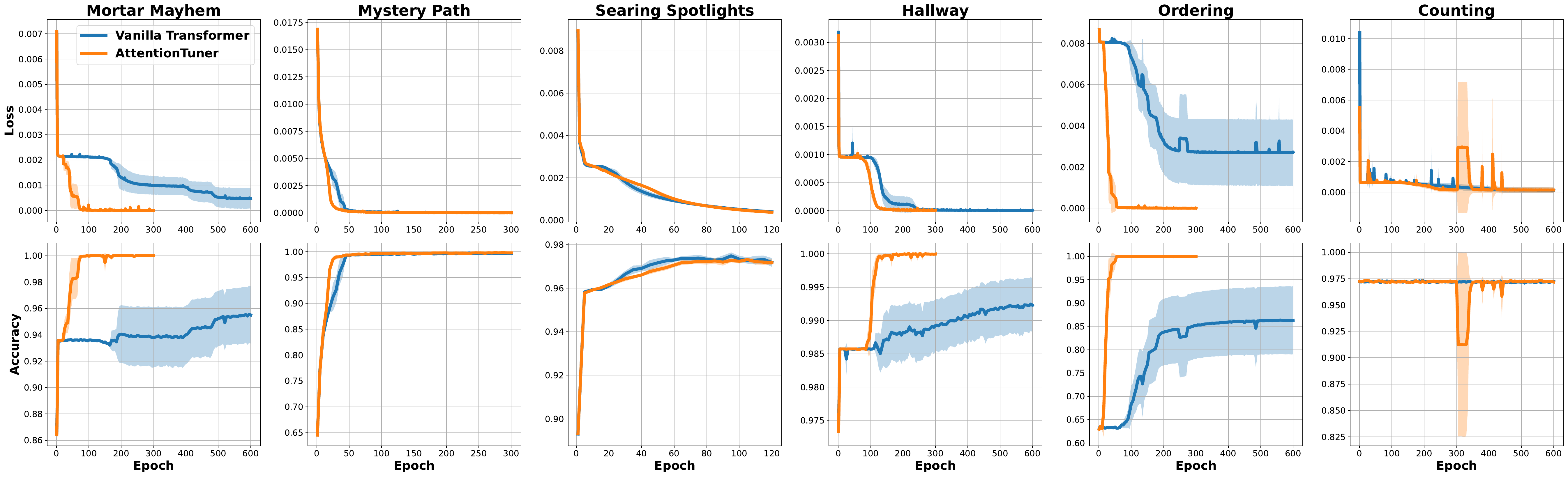}}
\caption{Mean learning curves with 90\% confidence intervals for Memory Gym tasks.}
\label{mean_curves}
\end{center}

\end{figure}

\section{Missing Annotations Ablation Data}
\label{sec:annotation}

\begin{table}[H]
\caption{Missing Annotations Ablations}
\label{tbl:annotations_ablation}
\begin{center}
\resizebox{\textwidth}{!}{
\begin{small}
\begin{sc}
\begin{tabular}{lcccccccr}
\toprule
Tasks & 0\% & 0.1\% & 1\% & 5\% & 10\% & 90\% & 100\% \\
\midrule
Mortar Mayhem & $20.8 \pm 42.2$ & $23 \pm 41.2$ & $80.3 \pm 41.3$ & $93.4 \pm 11.1$ & $100 \pm 0$ & $100 \pm 0$ & $99.8 \pm 0.4$ \\
Hallway & $42 \pm 50.5$ & $79.7 \pm 40$ & $96.7 \pm 6.8$ & $99.9 \pm 0.2$ & $99.8 \pm 0.5$ & $100 \pm 0$ & $99.9 \pm 0.1$ \\
\bottomrule
\end{tabular}
\end{sc}
\end{small}
}
\end{center}
\caption*{The table presents the success rates and their corresponding 90\% confidence intervals for tasks under different levels of missing annotations. The percentages indicate the proportion of demonstration trajectories with fully annotated memory dependency pairs. Figure~\ref{annotations_ablation} illustrates these results in a bar plot.}

\end{table}

\section{Imprecise Annotations Ablations Data}
\label{sec:imprecise}

\begin{table}[H]
\caption{Partially Imprecise Annotations Ablations}
\label{tbl:imprecise_annotations_ablation}
\begin{center}
\resizebox{\textwidth}{!}{
\begin{small}
\begin{sc}
\begin{tabular}{lcccccc}
\toprule
Tasks & 0 & 1 & 2 & 5 & 10 & 20 \\
\midrule
Mortar Mayhem & $99.8 \pm 0.4$ & $99.22 \pm 0.7$ & $99.7 \pm 0.5$ & $100 \pm 0.1$ & $79.3 \pm 38.2$ & $21.2 \pm 42$ \\
Hallway & $99.9 \pm 0.1$ & $99.9 \pm 0.1$ & $82 \pm 38.5$ & $81.7 \pm 38.8$ & $88.4 \pm 23.7$ & $56.7 \pm 45.6$ \\
\bottomrule
\end{tabular}
\end{sc}
\end{small}
}
\end{center}
\caption*{The table presents success rates and their corresponding 90\% confidence intervals for tasks with varying levels of imprecise annotations. For each memory association pair $(p, q)$, the recalled timestep $p$ is perturbed by a delta $\Delta$ drawn from a normal distribution $\mathcal{N}(0, \sigma)$. The top row indicates the standard deviation $\sigma$.}

\end{table}

\begin{table}[H]
\caption{Imprecise Annotations Ablations}
\label{tbl:imprecise_annotations_ablation2}
\begin{center}
\resizebox{\textwidth}{!}{
\begin{small}
\begin{sc}
\begin{tabular}{lcccccc}
\toprule
Tasks & 0 & $0.5$ & $0.75$ & 1 & $1.5$ & 2 \\
\midrule
Mortar Mayhem & $99.8 \pm 0.4$ & $100 \pm 0$ & $99.4 \pm 1$ & $100 \pm 0$ & $57.5 \pm 50.2$ & $20 \pm 42.6$ \\
Hallway & $99.9 \pm 0.1$ & $100 \pm 0$ & $75.9 \pm 36.4$ & $20.3 \pm 24.2$ & $24.5 \pm 36$ & $3.5 \pm 2.6$ \\
\bottomrule
\end{tabular}
\end{sc}
\end{small}
}
\end{center}
\caption*{The table presents success rates and their corresponding 90\% confidence intervals for tasks with varying levels of imprecise annotations. For each memory association pair $(p, q)$, the recalled timestep $p$ and the timestep of recall $q$ are both perturbed by deltas $\Delta_p$ and $\Delta_q$ drawn from a normal distribution $\mathcal{N}(0, \sigma)$. The top row indicates the standard deviation $\sigma$.}
\end{table}

\section{Annotating Memory Dependency Pairs}
\label{sec:annotate}

\begin{figure}[H]
\begin{center}
\centerline{\includegraphics[width=\textwidth, height=0.9\textheight, keepaspectratio]{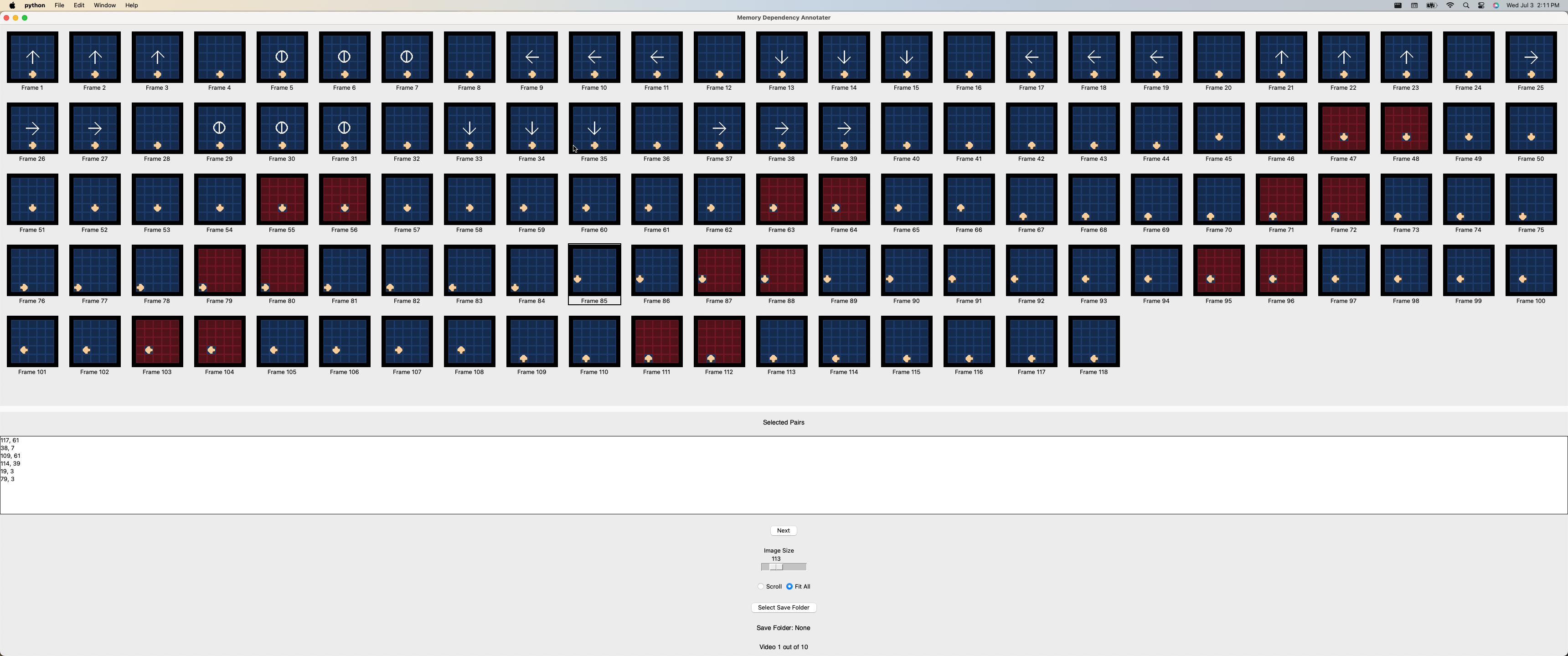}}
\caption{Graphical User Interface used to collect memory dependency pairs.}
\label{fig:gui}
\end{center}
\end{figure}

\begin{table}[H]
\centering
\begin{tabular}{lcc}
\toprule
\textbf{Task}           & \textbf{Demonstration Collection Time} & \textbf{Annotation Time} \\
\midrule
Mortar Mayhem           & 8 min 20 sec                     & 21 min 6 sec                           \\
Hallway                 & 2 min 9 sec                     & 5 min 59 sec                           \\
\bottomrule
\end{tabular}
\caption{The time it took to collect and annotate 10 demonstrations is recorded.}
\label{tbl:time}
\end{table}

Figure~\ref{fig:gui} shows the graphical user interface used to annotate demonstrations with memory dependency pairs. A single demonstration is loaded onto the GUI at a time. To annotate a memory dependency pair $(p, q)$, the annotator clicks on frame $p$ and then clicks on frame $q$. Table~\ref{tbl:time} compares the time it takes for humans to collect demonstrations versus annotating them.

\section{Computing the Value of Memory Dependency Pairs}
\label{sec:math}

\subsection{Mortar Mayhem}

There are 4000 demonstrations in the standard Mortar Mayhem training dataset (Appendix~\ref{settings}). Only 400 of these demonstrations have to be annotated to achieve a 100\% success rate (Figure~\ref{annotations_ablation}). It takes the vanilla Transformer 16,000 demonstrations to achieve a 100\% success rate (Figure~\ref{fig:scaling}). This means that each memory dependency pair annotation is worth $\frac{16000}{400} = 40$ additional demonstrations. According to Table~\ref{tbl:time}, annotating a single demonstration takes $\frac{1266}{10} = 126.6$ seconds while collecting 40 additional demonstration takes $\frac{500}{10} * 40 = 2000$ seconds. This results in a human labor time savings factor of $\frac{2000}{126.6} = 15.7977883 \approx 16$.

\subsection{Hallway}

There are 5000 demonstrations in the standard Hallway training dataset (Appendix~\ref{settings}). Only 250 of these demonstrations have to be annotated to achieve a 100\% success rate (Figure~\ref{annotations_ablation}). It takes the vanilla Transformer 10,000 demonstrations to achieve a 100\% success rate (Figure~\ref{fig:scaling}). This means that each memory dependency pair annotation is worth $\frac{10000}{250} = 40$ additional demonstrations. According to Table~\ref{tbl:time}, annotating a single demonstration takes $\frac{359}{10} = 35.9$ seconds while collecting 40 additional demonstrations takes $\frac{129}{10} *  40 = 516$ seconds. This results in a human labor time savings factor of $\frac{516}{35.9} = 14.3732591 \approx 14$.

\section{Expert Truth Attention Heatmaps}
\label{expert}

\begin{figure}[H]
\begin{center}
\centerline{\includegraphics[width=\textwidth, height=0.9\textheight, keepaspectratio]{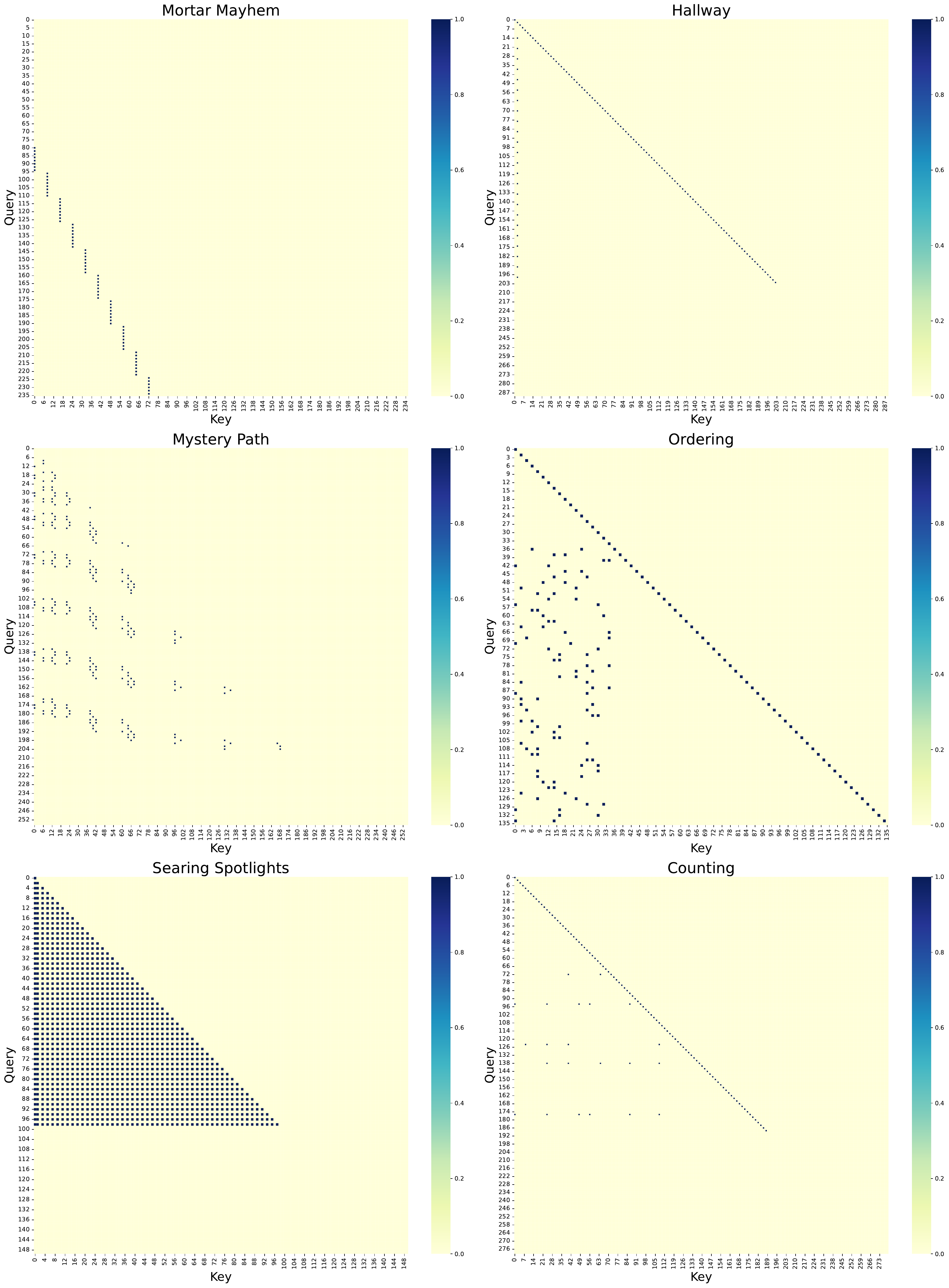}}
\caption{Sample expert attention heatmaps for $E$}
\label{expert_heatmaps}
\end{center}

\end{figure}

\begin{figure}[H]
\begin{center}
\centerline{\includegraphics[width=\textwidth]{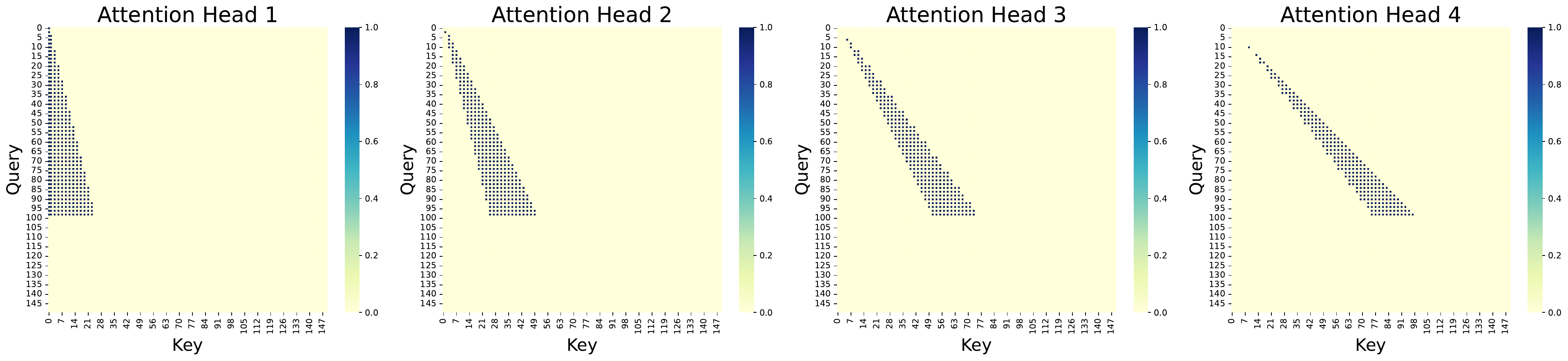}}
\caption{Sample expert attention heatmaps for split memory loss on Searing Spotlights. Notably, this experiment uniquely employs 4 self-attention heads, diverging from the typical configuration of 2 self-attention heads used in all other experiments.}
\label{split}
\end{center}

\end{figure}

\section{Learned Heatmaps}
\label{learned_heatmaps}

\begin{figure}[H]
\begin{center}
\centerline{\includegraphics[width=\textwidth, height=0.9\textheight, keepaspectratio]{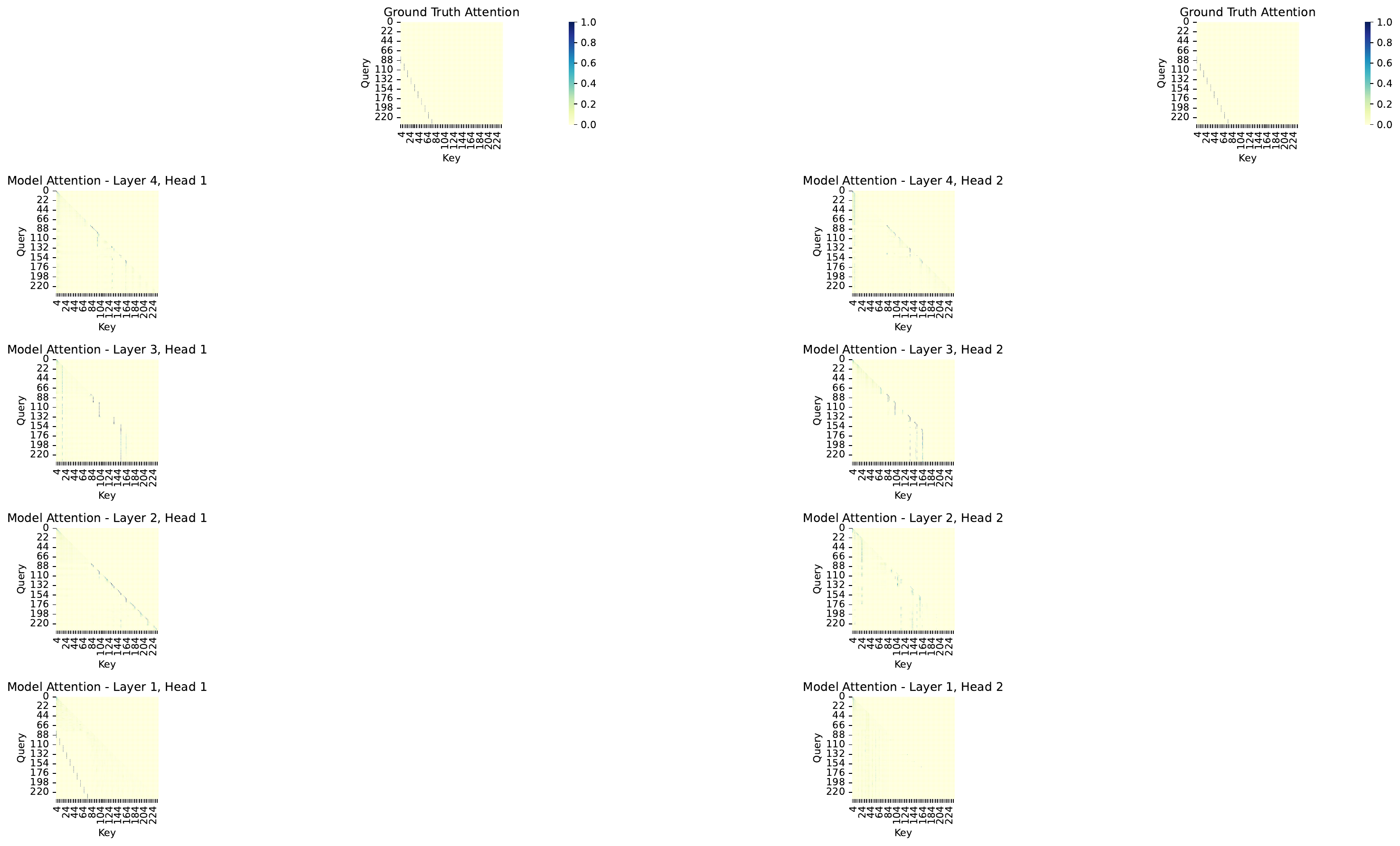}}
\caption{AttentionTuner's learned self-attention heatmap for all attention heads on Mortar Mayhem.}
\label{mem_heatmap}
\end{center}

\end{figure}

\begin{figure}[H]
\begin{center}
\centerline{\includegraphics[width=\textwidth, height=0.9\textheight, keepaspectratio]{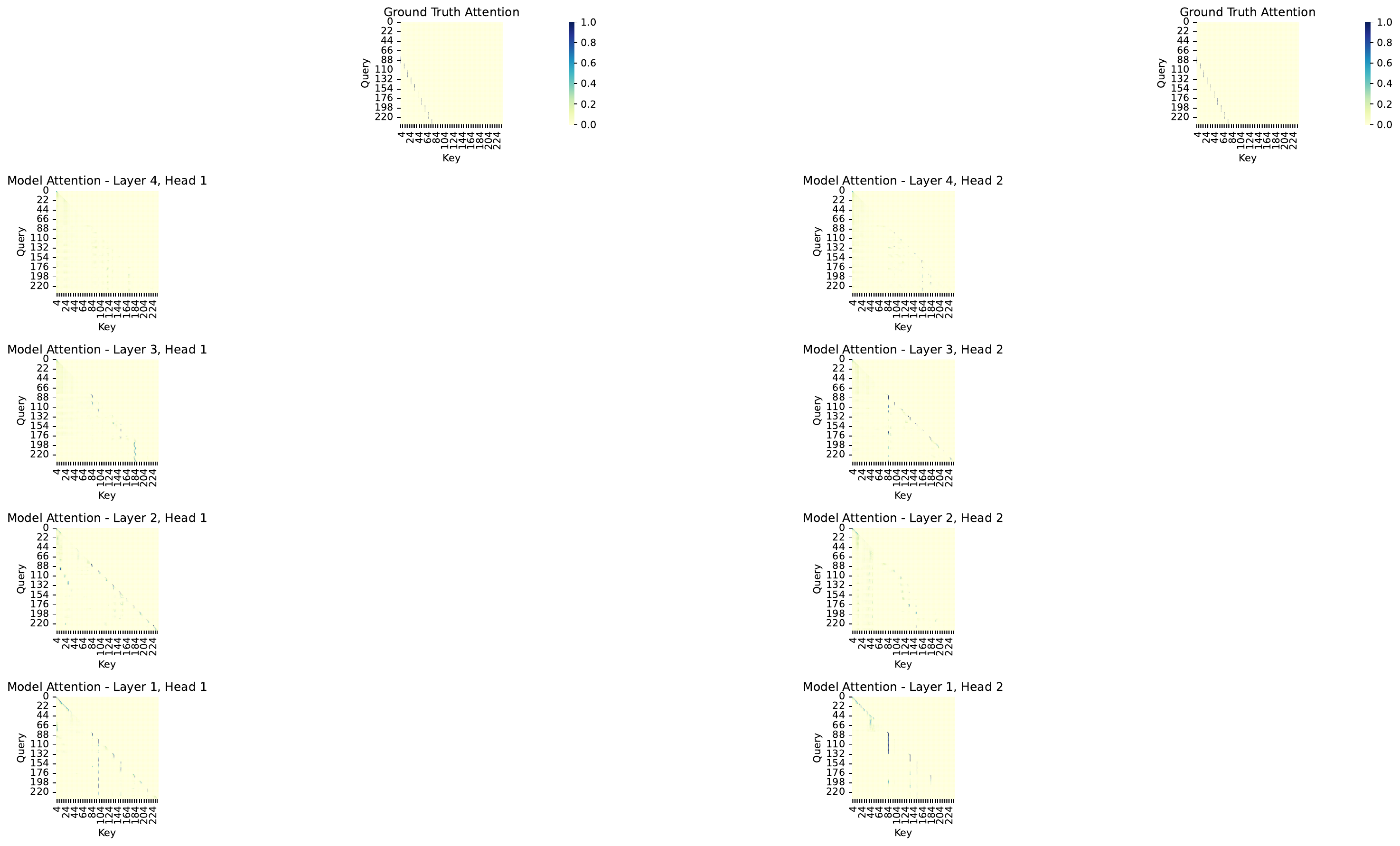}}
\caption{Vanilla Transformer's learned self-attention heatmap for all attention heads on Mortar Mayhem.}
\label{norm_heatmap}
\end{center}

\end{figure}

Figures \ref{mem_heatmap} and \ref{norm_heatmap} illustrate all self-attention heads in the Transformer as learned by AttentionTuner and the vanilla Transformer after training on Mortar Mayhem. Figure \ref{mem_heatmap} demonstrates that, with the application of memory loss (Equation~\ref{memloss}) to the initial head of the first layer, the correct memory mechanism was effectively acquired. Conversely, Figure \ref{norm_heatmap} indicates that in the absence of memory loss, solely employing the behavioral cloning objective did not facilitate the acquisition of the precise memory mechanism. However, a partially accurate memory mechanism is observable in the first head of the second layer.

\section{Computational Resources}
\label{compute}

All experiments were conducted on Nvidia A40 and A100 GPUs with 40 or 80 GB of memory. The computational node featured two Intel Xeon Gold 6342 2.80GHz CPUs with 500 GB RAM. Experiment durations varied between 1 and 4.5 hours.

\end{document}